\documentclass[runningheads]{llncs}
\usepackage{graphicx}

\usepackage{tikz}
\usepackage{comment}
\usepackage{amsmath,amssymb} 
\usepackage{color}
\usepackage{pifont}
\newcommand{\cmark}{\ding{51}}%
\newcommand{\xmark}{\ding{55}}%

\usepackage[accsupp]{axessibility}  

\usepackage{kotex}

\usepackage{multirow, makecell} 
\usepackage{booktabs} 

\newcommand{\ourmodel}{our model }
\newcommand{\hairloss}{local-style-matching loss }

\usepackage{hyperref} 

\begin{document}
\pagestyle{headings}
\mainmatter
\def\ECCVSubNumber{6395}  

\title{Style Your Hair: Latent Optimization for Pose-Invariant Hairstyle
Transfer via Local-Style-Aware Hair Alignment} 


\titlerunning{Style Your Hair}
\author{Taewoo Kim\inst{*} \and
Chaeyeon Chung\inst{*} \and
Yoonseo Kim\inst{*} \and \\
Sunghyun Park \and
Kangyeol Kim \and
Jaegul Choo
}
%
\authorrunning{T. Kim et al.}
%
\institute{Korea Advanced Institute of Science and Technology, Daejeon, South Korea \\
\email{\{specia1ktu, cy\_chung, grandchasevs\\psh01087, kangyeolk, jchoo\}@kaist.ac.kr}\\
* indicates equal contributions.}
\maketitle

\vspace{-0.5cm}
\begin{abstract}
Editing hairstyle is unique and challenging due to the complexity and delicacy of hairstyle.
Although recent approaches significantly improved the hair details, these models often produce undesirable outputs when a pose of a source image is considerably different from that of a target hair image, limiting their real-world applications.
HairFIT, a pose-invariant hairstyle transfer model, alleviates this limitation yet still shows unsatisfactory quality in preserving delicate hair textures.
To solve these limitations, we propose a high-performing pose-invariant hairstyle transfer model equipped with latent optimization and a newly presented local-style-matching loss.
In the StyleGAN2 latent space, we first explore a pose-aligned latent code of a target hair with the detailed textures preserved based on local style matching.
Then, our model inpaints the occlusions of the source considering the aligned target hair and blends both images to produce a final output.
The experimental results demonstrate that our model has strengths in transferring a hairstyle under larger pose differences and preserving local hairstyle textures. The codes are available at \href{https://github.com/Taeu/Style-Your-Hair}{https://github.com/Taeu/Style-Your-Hair}.
\end{abstract}
\vspace{-0.7cm}

\begin{figure}
    \centering
    \includegraphics[width=\linewidth]{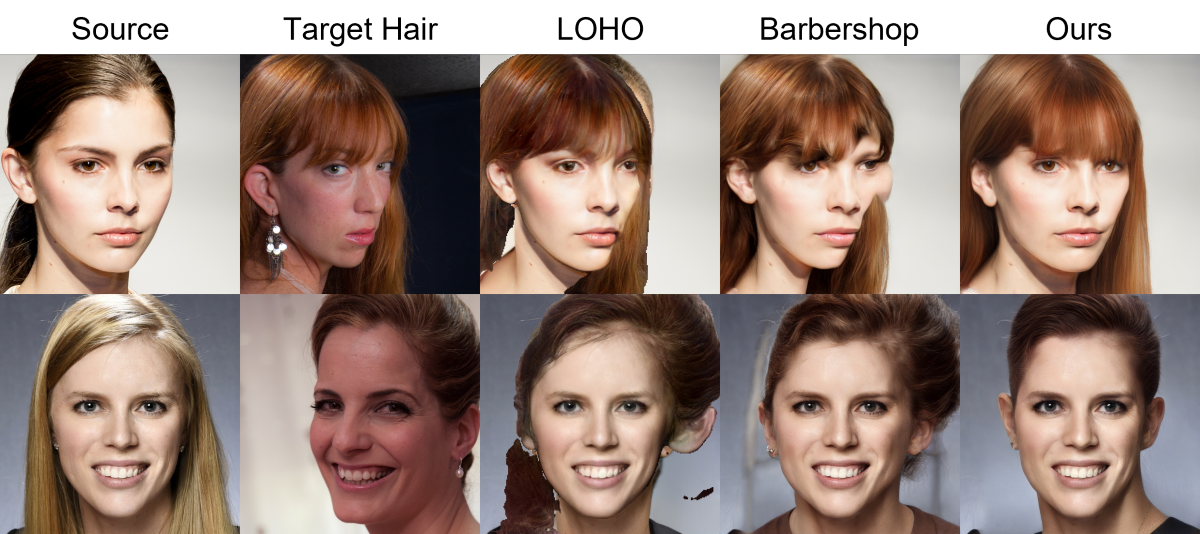}
    \caption{Our model produces more realistic results compared to LOHO~\cite{saha2021loho} and Barbershop~\cite{zhu2021barbershop} even with a large pose difference between a source and a target hair.}
    \label{fig:teaser}
\end{figure}

\section{Introduction}
With the advance of conditional generative adversarial networks (GANs)~\cite{goodfellow2014generative,odena2017conditional,jo2019sc}, editing facial attributes has drawn great attention and shows a promising result on editing multiple attributes.
Despite the success, modifying strongly correlated facial attributes is still challenging, often beyond the capacity of existing editing models.
In this paper, we focus on hairstyle editing, which aims at transferring a target hairstyle to a source image, proposing high-performance neural networks to solve the problem.
Hairstyle editing is similar to that of a facial attribute, but it has unique, challenging aspects: (1) Due to the hairstyle's complexity and delicacy,  preserving its strands given an arbitrary hairstyle is highly demanding. (2) Transferred hairstyle requires to be exactly fitted to a given source image. 
These challenges make the previous approaches for editing the specified facial attributes less suitable for this problem.

Recent solutions for hairstyle transfer address the problem with the power of a pre-trained image generator.
For example, LOHO~\cite{saha2021loho} and Barbershop~\cite{zhu2021barbershop} largely enhance the visual quality of the generated images via latent optimization based on StyleGAN2~\cite{karras2020stylegan2}.
However, these approaches produce undesirable outputs (See Fig.~\ref{fig:teaser}) when handling a target and source image pair with a significant pose difference.

To the best of our knowledge, HairFIT~\cite{chung2021hairfit} is the only work to address the pose difference issue between a source and target image. HairFIT presents a pose-invariant hairstyle transfer model where a target hairstyle is aligned to a source image pose using a flow-based warping module trained on multi-view datasets such as VoxCeleb~\cite{nagrani2017voxceleb} and K-hairstyle dataset~\cite{kim2021k}.
Although its attempt, HairFIT requires a high-quality multi-view hairstyle dataset during training, and it falls behind state-of-the-art models~\cite{saha2021loho,zhu2021barbershop} in light of hair preserving capacity.

In response to these limitations, we present a framework that performs a \textit{high-quality} pose-invariant hairstyle transfer based on latent optimization \textit{without multi-view dataset}.
Specifically, given a source and a target hair image, our model generates a hair-transferred output through embedding, hair pose alignment, inpainting, and blending step.
We first take advantage of GAN-inversion algorithms~\cite{abdal2019image2stylegan,abdal2020image2stylegan++,zhu2020ii2s}, feeding a source and a target hair image, for the purpose of obtaining latent codes residing in the StyleGAN2 space~\cite{karras2020stylegan2}, respectively.
Next, we navigate the StyleGAN2 space to optimize the latent code of the target hair image to follow a source image pose.
During the pose alignment, we utilize a newly-presented \hairloss to penalize visually degraded hair textures by locally comparing the original target hair with the aligned one.
In the inpainting, we first obtain a segmentation mask to guide the latent code of the source to fill the occluded regions by its hair.
We optimize the latent code of the source image to follow the obtained segmentation mask.
Lastly, we blend the aligned target hairstyle and the inpainted source image with a final optimization step.
In this manner, \ourmodel is able to transfer a target hairstyle to a source image overcoming the difference in poses as well as successfully preserving the fine details of the target hair.
Experiments demonstrate the superiority of our model in a quantitative and qualitative manner.
Our contributions are summarized as follows:
\begin{itemize}
    \item We propose a framework that achieves a \textit{high-quality} pose-invariant hairstyle transfer based on latent optimization \textit{without multi-view dataset}.
     \item We present \hairloss to maintain the fine details of a hairstyle during pose optimization.
    \item Our model achieves state-of-the-art performance in quantitative and qualitative evaluations with various datasets. 
\end{itemize}

\section{Related Work}

\subsection{Latent space manipulation}
With the understanding of the latent space in GANs, recent approaches based on latent space manipulation~\cite{harkonen2020ganspace,shen2020interpreting,viazovetskyi2020distillation} have shown promising results in image editing.
For example, GANSpace~\cite{harkonen2020ganspace} and InterfaceGAN~\cite{shen2020interpreting} modify facial attributes via manipulation in the latent space of StyleGAN~\cite{karras2019stylegan}.
While the former takes advantage of principal component analysis, the latter utilizes semantic scores to identify disentangled directions related to the target attributes.
In a similar manner, Viazovetskyi et al.~\cite{viazovetskyi2020distillation} and Zhuang et al.~\cite{zhuang2021enjoy} attempt to edit the images by shifting latent vectors to semantically meaningful directions in the latent space of StyleGAN2~\cite{karras2020stylegan2}, which can easily be obtained by a pre-trained face classifier or learned transformations.

Recent hairstyle transfer approaches also actively utilize latent space manipulation to synthesize high-quality images.
LOHO~\cite{saha2021loho} and Barbershop~\cite{zhu2021barbershop} edit hairstyles by manipulating the extended StyleGAN2 latent space~\cite{zhu2020ii2s} via latent optimization.
These methods not only significantly enhance the visual quality of the generated images but also preserve the semantic details of the target images.
In particular, Barbershop introduces $FS$ space with a larger capacity than the original StyleGAN2 latent space, where the original hair structure is well-preserved.
In this work, we leverage latent optimization to reach the photo-realistic image quality.
Our model mainly focuses on the pose alignment of a target hairstyle to a source image without losing its detailed hair texture based on \hairloss to achieve a pose-invariant hairstyle transfer.

\subsection{Hairstyle Transfer}


GAN-based facial image editing~\cite{jiang2020psgan,jo2019sc,lee2020maskgan,portenier2018faceshop,xiao2021sketchhairsalon,yang2020deep} successfully modifies the target facial attributes such as a facial expression or makeup style while maintaining other features.
Common approaches for facial image editing are to utilize hand-drawn sketches~\cite{jo2019sc,portenier2018faceshop,xiao2021sketchhairsalon,yang2020deep} or user-edited semantic masks\cite{lee2020maskgan} as the conditions to precisely guide the manipulated appearance.


In spite of the remarkable progress in facial image editing, hairstyle transfer is still tricky, considering the diversity and intricacy of hairstyles.
In practice, a hairstyle transfer is required to convey a wide range of target hairstyles to a given source image while preserving their subtle hair strands and color.
As a prior work, MichiGAN~\cite{tan2020michigan} presents a hairstyle transfer framework aiming to preserve the detailed textures of a target hairstyle.
Specifically, MichiGAN leverages different conditional generators responsible for decomposed hairstyle attributes (\textit{i.e.}, hair shape, and appearance).
Moreover, LOHO~\cite{saha2021loho} achieves visually pleasant image quality through latent optimization and hair-related losses for reflecting a target hairstyle features.
Barbershop~\cite{zhu2021barbershop} also proposes a latent optimization approach and further improves the visual quality of the outputs based on the newly presented $FS$ space.
Barbershop utilizes $F$ tensor in $FS$ space to enhance the capability of preserving the overall structure of a target image, including delicate hair structure.

However, since the existing approaches have been developed to handle the images, where the head poses of a source and a target are aligned, they show limited generalization capacity for dealing with the inputs having a large pose difference.
To tackle this problem, HairFIT~\cite{chung2021hairfit} introduces a pose-invariant hairstyle transfer with flow-based target hair warping and semantic-region-aware inpainting.
HairFIT leverages an optical flow estimation network and a multi-view hairstyle dataset~\cite{kim2021k} to align the target hair to the source face.
Despite the aid of the high-quality multi-view dataset, the model fails to preserve the detailed features of hairstyles comprehensively.
In this paper, we propose a novel latent optimization framework for pose-invariant hairstyle transfer to synthesize high-quality images regardless of the pose differences.


\section{Method}

\subsection{Overview}
\begin{figure}[t!]
    \centering
    \includegraphics[width=1.0\linewidth]{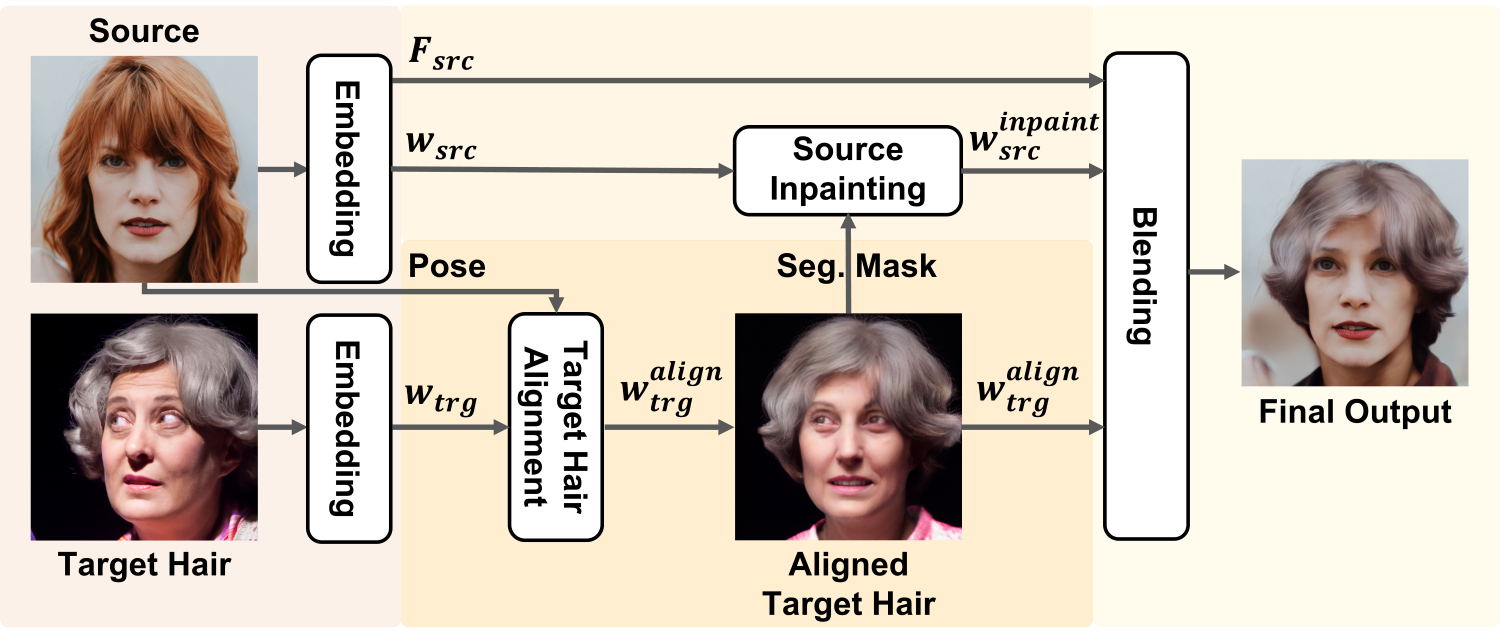}
    \caption{An overview of our framework. First, we obtain $w_{src}$, $w_{trg}$, and $F_{src}$ by embedding a source and a target hair image into $W+$ and $FS$ space. Then, we optimize $w_{trg}$ to follow the source pose, resulting in $w^{align}_{trg}$. With the segmentation mask of aligned target hair, we find $w^{inpaint}_{src}$, where the source occlusions are inpainted. Finally, we blend $F_{src}$, $w^{inpaint}_{src}$, and $w^{align}_{trg}$ to generate the final output.}
    \label{fig:overview}
\end{figure}

Our framework consists of several optimization steps described in Fig.~\ref{fig:overview}.
We first find latent codes $w \in \mathbb{R}^{18 \times 512}$ of a source image $\mathbf{I}_{src} \in \mathbb{R}^{C \times H \times W}$ and a target hairstyle image $\mathbf{I}_{trg}$ in $W$+ space using the existing GAN-inversion algorithms~\cite{zhu2021barbershop,zhu2020ii2s}.
Then, we optimize the target hair latent codes to have the pose aligned to $\mathbf{I}_{src}$.
While aligning the pose, we mainly focus on preserving fine details of the target hair with a newly-presented local-style-matching loss.
Local-style-matching loss allows preserving each local texture in the aligned target hair by matching the corresponding region of a similar style from the original target hair.
For the next step, we inpaint the source regions occluded by its original hair by optimizing the source latent codes.
Lastly, we blend the aligned target hairstyle and the inpainted source image for the final output.

\subsection{Embedding}
First of all, we obtain the latent codes of each reference image (\textit{i.e.}, source and target images) before pose alignment and blending.
Given a source image $\mathbf{I}_{src}$ and a target hair image $\mathbf{I}_{trg}$, we find the source latent codes $w_{src}$ and the target latent codes $w_{trg}$ in an extended latent space of StyleGAN2 denoted as $W+$ space~\cite{abdal2019i2s}.
We employ an improved embedding algorithm~\cite{zhu2020ii2s} to enhance the reconstruction and editing quality.
Moreover, we embed $\mathbf{I}_{src}$ to $FS$ space following Barbershop~\cite{zhu2021barbershop} to gain $\mathbf{F}_{src} \in \mathbb{R}^{32 \times 32 \times 512}$, which preserves the detailed structure of the source image by encoding the spatial information.

\subsection{Target Hair Alignment}
\begin{figure}[t!]
    \centering
    \includegraphics[width=1.0\linewidth]{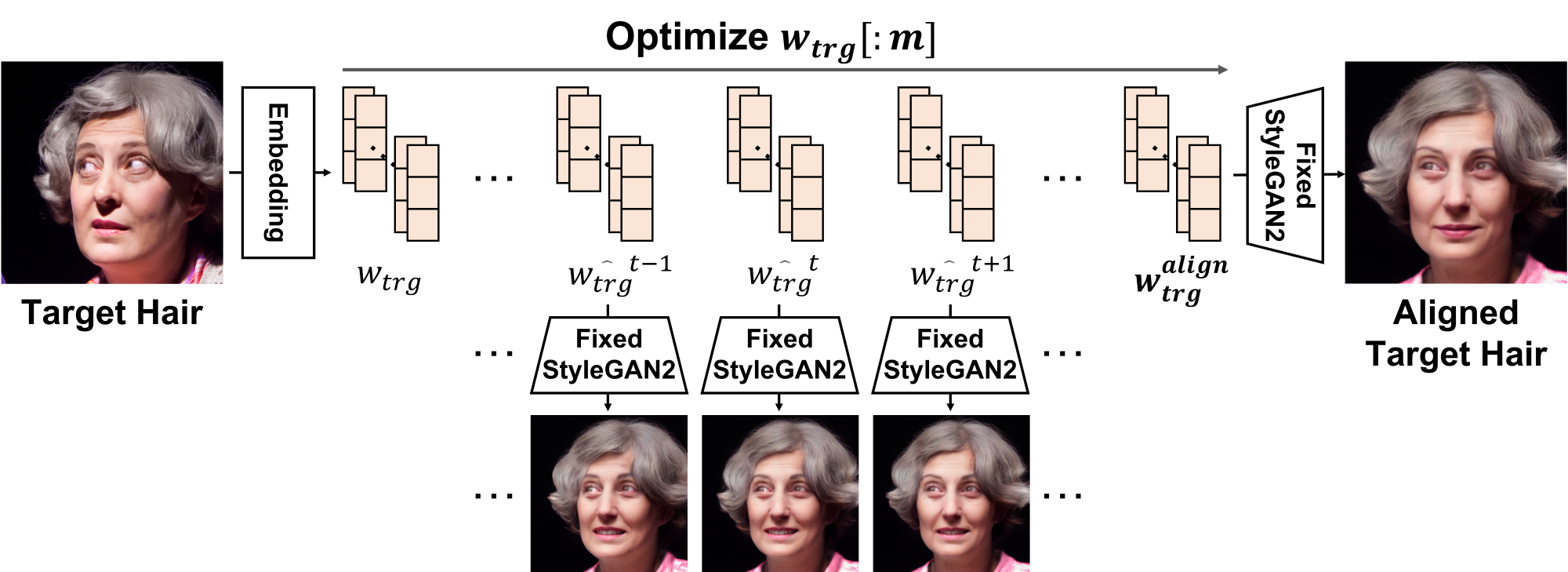}
    \caption{Target hair alignment. We obtain the aligned target hair latent codes $w^{align}_{trg}$ by optimizing the first $m$ vectors of $w_{trg}$ to have a source pose with its hairstyle preserved.}
    \label{fig:align}
\end{figure}

To transfer the hairstyle regardless of the pose differences, we align the target hairstyle to the source face via the latent optimization, as presented in Fig.~\ref{fig:align}.
Starting from $w_{trg}$, we aim to find $w^{align}_{trg}$, where the head pose and face shape are aligned to $\mathbf{I}_{src}$, while other features, especially the hairstyle, correspond to $\mathbf{I}_{trg}$.
We optimize the first $m$ style vectors among 18 style vectors of $w_{trg}$ to optimize coarse style vectors rather than fine style vectors~\cite{karras2019stylegan}.
We set $m$ as 6 in our experiments.

\noindent\textbf{Pose Align Loss.}
To modify $w_{trg}$ to have a source pose, we propose a novel pose align loss $\mathcal{L}_{pose}$ based on 3D facial keypoints.
Since the hairstyle significantly depends on other facial features (\textit{e.g.}, face shape and location of eyes), the head pose alone is insufficient to fully guide the target hair alignment.
Thus, we leverage 3D facial keypoints, which effectively represent the overall facial features as well as the head pose.
With the source 3D facial keypoints, we can provide detailed supervision of which shape and pose ${w_{trg}}$ should pursue.

$\mathcal{L}_{pose}$ computes the L2 distance between the 3D keypoint heatmaps of $\mathbf{I}_{src}$ and the aligned target hair image as:
\begin{equation}
    \mathcal{L}_{pose} = \frac{1}{N_H}\lVert \mathbf{H}_{src} - E(G(\hat{w_{trg}})) \rVert^{2}_{2}.
\end{equation}
$N_H$ indicates the number of elements in a 3D keypoint heatmap $\mathbf{H} \in \mathbb{R}^{68 \times H \times W}$ and $\mathbf{H}_{src}=E(\mathbf{I}_{src})$, where $E$ is a pre-trained keypoint extractor~\cite{bulat2017far}. $G$ is a pre-trained StyleGAN2 generator and $\hat{w_{trg}}$ indicates the optimized $w_{trg}$ in progress.

\noindent\textbf{Local-Style-Matching Loss.}
To preserve locally distinct hairstyles, we newly present a local-style-matching loss, which matches similar local styles between the target hair and the aligned target hair.
Basically, we utilize a style loss based on the Gram matrix~\cite{gatys2016image}, which captures the repeated patterns (\textit{i.e.}, texture) of given features.
A style loss $\mathcal{L}_{style}$ measures the L2 distance between the Gram matrix of feature maps extracted by a $\mathrm{VGG}$ network~\cite{simonyan2014very}, formulated as:
\begin{equation}
   \mathcal{L}_{style}(\cdot,\cdot) = \frac{1}{V}\sum_{i=1}^V \frac{1}{N_{\mathcal{G}^i}} \lVert \mathcal{G}^i(\mathrm{VGG}^i(\cdot)) - \mathcal{G}^i(\mathrm{VGG}^i(\cdot))\rVert^{2}_{2},
\end{equation}
where $V$ indicates the number of $\mathrm{VGG}$ layers we use, which are $relu1\_2$, $relu2\_2$, $relu3\_3$, and $relu4\_3$ layer of $\mathrm{VGG}$~\cite{chung2021hairfit,saha2021loho,zhu2021barbershop}.
Also, $N_{\mathcal{G}^i}$ represents the number of elements in $\mathcal{G}^{i}$.
Here, $\mathcal{G}^{i}$ and $\mathrm{VGG}^{i}$ indicate the $i$-th Gram matrix and $i$-th layer of $\mathrm{VGG}$, respectively.
$\mathcal{G}^{i}$ is calculated as ${v^{i}}^{\intercal}v^{i}$, where $v^{i} \in \mathbb{R}^{H^{i}W^{i} \times N_{C^{i}}}$ corresponds to the activation of $\mathrm{VGG}^{i}$. 

\begin{figure}[t!]
    \centering
    \includegraphics[width=1.0\linewidth]{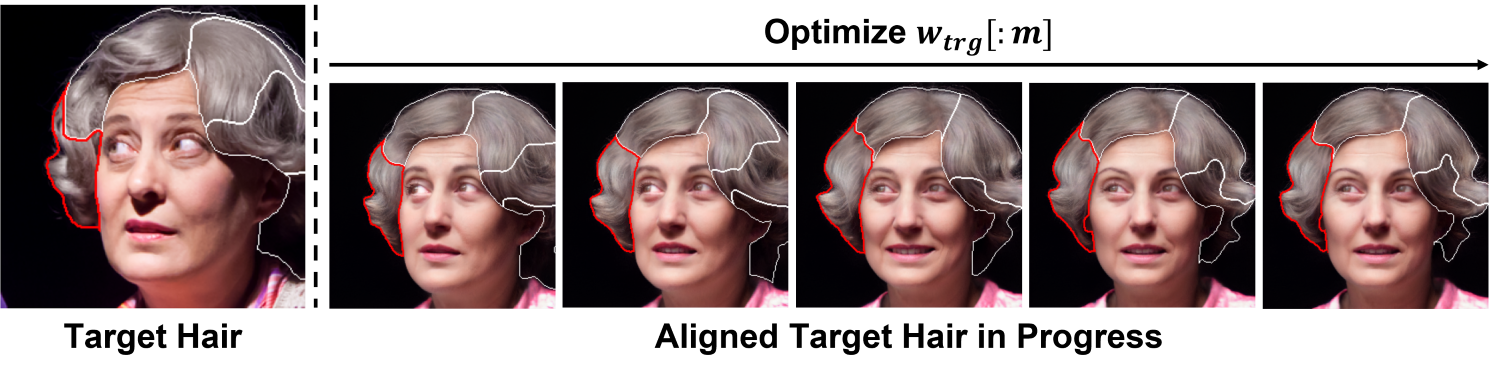}
    \caption{Local-style-matching loss. During target hair alignment, a local-style-matching loss is applied to style regions in the target hair and those in the aligned target hair. The white boundary regions are segmented style regions, and the red boundary regions describe an example of a consistently tracked style region.}
    \label{fig:lsm}
\end{figure}

In local-style-matching loss $\mathcal{L}^{LSM}_{style}$, we first identify \textit{style regions} each of which includes locally different style and apply $L_{style}$ to each style region, respectively.
To identify the style regions, we leverage a simple linear iterative clustering (SLIC)~\cite{achanta2012slic}.
The SLIC is an algorithm that conducts a K-means clustering based on the similarity of color and spatial distance between pixels.
Since the SLIC considers both the appearance and location of neighboring pixels, it can successfully segment the target hair into proper style regions.

As presented in the first column of Fig.~\ref{fig:lsm}, we first find the style regions in the hair of $\mathbf{I}_{trg}$.
Then, during the latent optimization, we detect the style regions of $G(\hat{w_{trg}})$ and match each region to the most similar style region of $\mathbf{I}_{trg}$, as shown in the rest columns of Fig.~\ref{fig:lsm}.
In each step, we track the regions of similar style by setting the same label to the region of the closest centroid compared to the previous step.
Fig.~\ref{fig:lsm} shows that an example style region marked with a red boundary is successfully tracked based on the proposed algorithm.
$SLIC_{hair}(\mathbf{I}) \in \{0,1\}^{N_{style} \times H \times W} $ indicates style region masks extracted from a hair region of $\mathbf{I}$ using the SLIC algorithm. Here, $N_{style}$ indicates the number style regions.
We set $N_{style}$ as $5$ in our experiments.
$\mathcal{L}^{LSM}_{style}$ is formulated as:
\begin{equation}
    \mathcal{L}^{LSM}_{style} = \sum_{i=1}^{N_{style}}\mathcal{L}_{style}(SLIC^i_{hair}(\mathbf{I}_{trg}) \odot \mathbf{I}_{trg} , SLIC^i_{hair}(G(\hat{w_{trg}})) \odot G(\hat{w_{trg}}) ).
\end{equation}
$SLIC^i_{hair}(\cdot)$ is the $i$-th channel of $SLIC_{hair}(\cdot)$ and $\odot$ indicates element-wise product.
Note that a valid region of each channel, where the style region mask corresponds to 1, is cropped before calculating the style loss.

\noindent\textbf{Regularization Loss.}
We add a step-wise regularization loss to keep the overall features of $\hat{w_{trg}}$, especially hairstyle, similar to the previous step.
The regularization loss $\mathcal{L}_{reg}$ encourages a stable optimization via a gradual modification without a noticeable loss of the original hairstyle features.
$\mathcal{L}_{reg}$ is formulated as:
\begin{equation}
    \mathcal{L}_{reg} = \frac{1}{N_w}\lVert \Delta \hat{w_{trg}} \rVert^{2}_{2},
\end{equation}
where $N_w$ indicates the number of elements in $w$.
$\Delta \hat{w_{trg}}$ at step $t$ is obtained by $\hat{w_{trg}}^{t} - \hat{w_{trg}}^{t-1}$, where $t$ ranges from 2 to the total number of steps.

Formally, the total objective function in the target hair alignment step is $\mathcal{L}_{pose} + \lambda^{LSM}_{style} \mathcal{L}^{LSM}_{style} + \lambda_{reg} \mathcal{L}_{reg}$, where $\lambda^{LSM}_{style}$ and $\lambda_{reg}$ denote the hyper-parameters to control relative importance between different losses.

\begin{figure}[t!]
    \centering
    \includegraphics[width=1.0\linewidth]{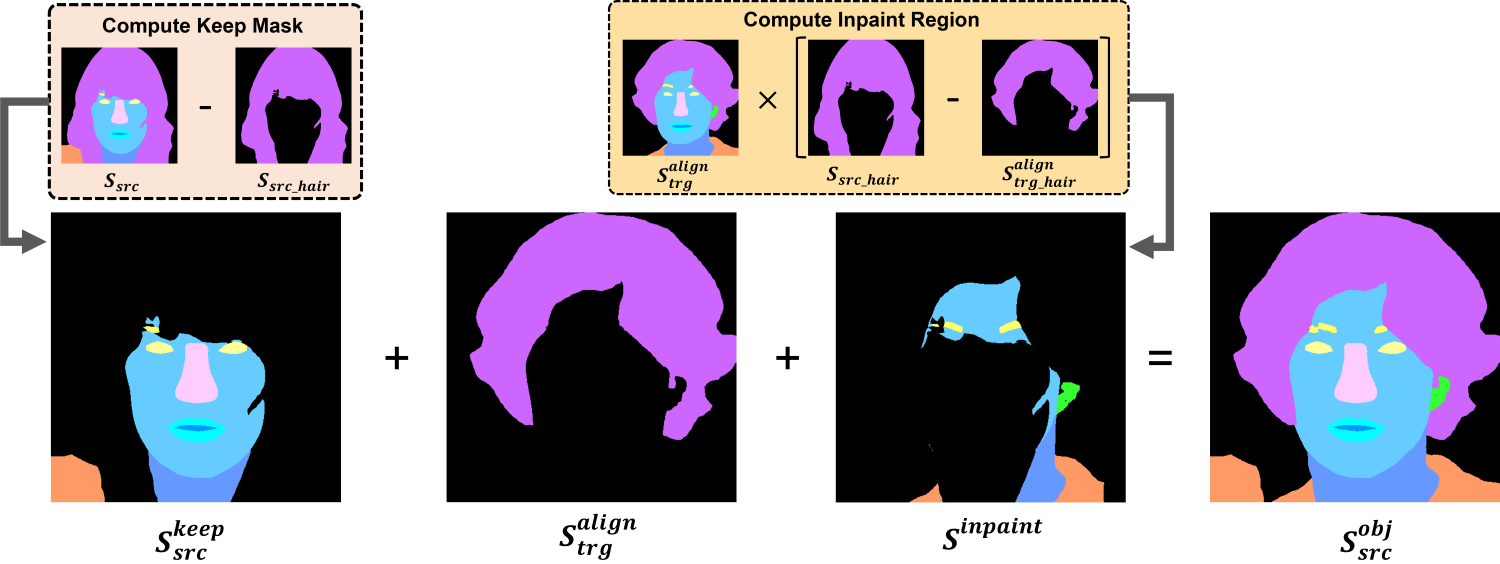}
    \caption{Generation of an objective label. For source inpainting, we create an objective label $\mathbf{S}^{obj}_{src}$ to guide the occluded regions to be inpainted with proper semantics.}
    \label{fig:align_mask}
\end{figure}

\subsection{Source Inpainting}
Source inpainting step aims to inpaint the regions occluded by the original source hair.
As shown in Fig.~\ref{fig:overview}, if we remove the source hair region from the source image, the occluded region should be filled with the proper semantics (\textit{e.g.}, forehead, face, neck, clothes, and background) to fit the aligned target hair.

To find the inpainted source latent code $w^{inpaint}_{src}$, we generate an objective label $\mathbf{S}^{obj}_{src} \in \mathbb{Z}^{H \times W}$ to guide the occluded regions to be filled with the appropriate semantic regions.
$\mathbf{S}^{obj}_{src}$ is generated by the following process, as also described in Fig.~\ref{fig:align_mask}. 
First, we compute a keep label $\mathbf{S}^{keep}_{src}$, which indicates the regions that need to be maintained in the source, by removing a source hair region $\mathbf{S}_{src\_hair}$ from a source semantic label $\mathbf{S}_{src}$.
Here, $\mathbf{S}_{src}$ is estimated by a pre-trained segmentation network~\cite{yu2018bisenet}.
Next, we calculate a label of regions to be inpainted $\mathbf{S}^{inpaint}$ as described in Fig.~\ref{fig:align_mask}.
Finally, we obtain $\mathbf{S}^{obj}_{src}$ which indicates the inpainting regions of the source image considering the aligned target hair.
Now, we optimize ${w}^{inpaint}_{src}$ to follow the given $\mathbf{S}^{obj}_{src}$.
Here, as in the target hair alignment step, we optimize the first $m$ $w$ vectors to newly generate coarse features to fill the occlusions while preserving the fine details or the overall appearance of the source.
For optimizing ${w}^{inpaint}_{src}$, we use a pixel-wise cross-entropy loss between the label of $\mathbf{S}^{obj}_{src}$ and a segmentation probability heatmap of the generated image, which consists of 16 semantic region categories.
The heatmap is estimated by the pre-trained segmentation network.

\subsection{Blending}
The final optimization step aims to find a blending weight $w^{weight}$ that merges the optimized latent codes from the previous steps to generate the final output.
\begin{figure}[t!]
    \centering
    \includegraphics[width=1.0\linewidth]{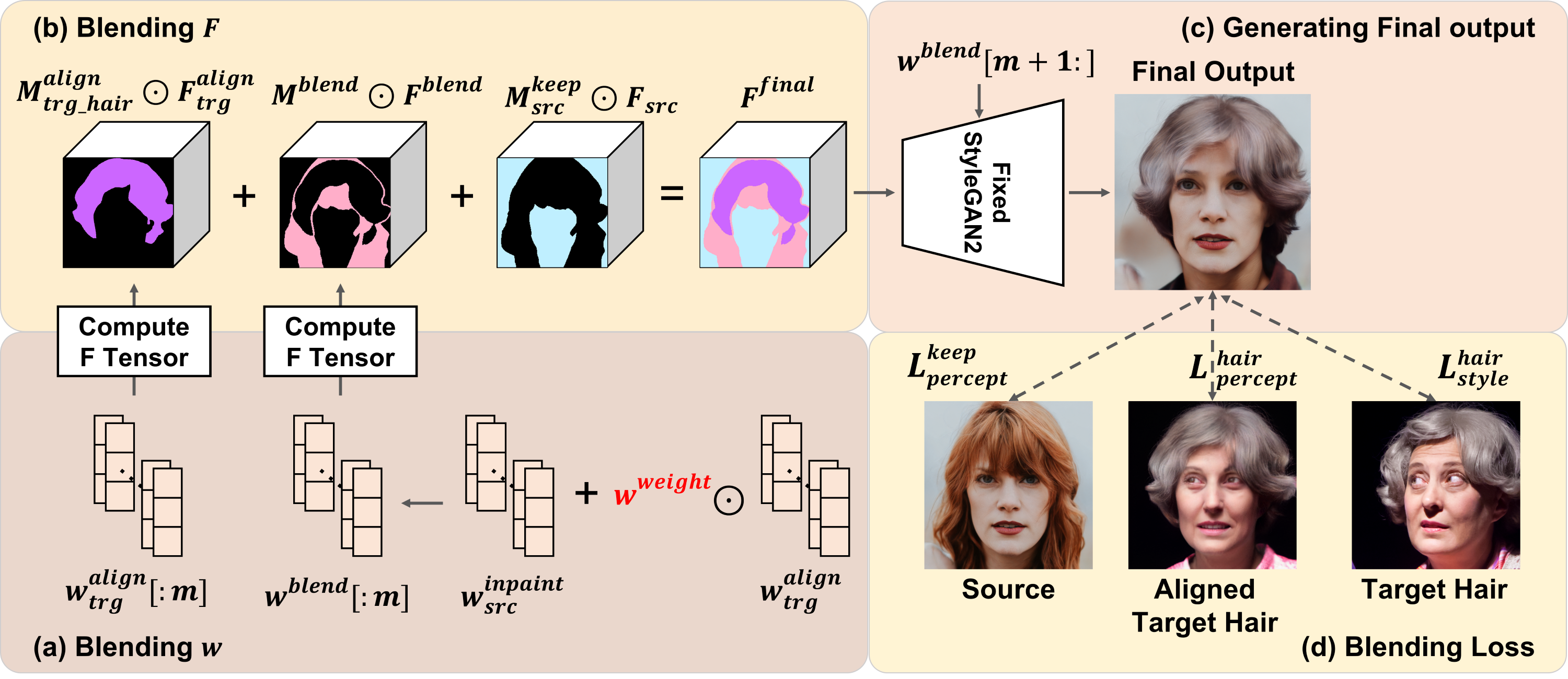}
    \caption{Blending. (a) $w$ vectors from the previous steps are blended with the optimized blending weight $w^{weight}$ to obtain $w^{blend}$. (b) Next, we combine $\mathbf{F}^{align}_{trg}$, $\mathbf{F}^{blend}$, and $\mathbf{F}_{src}$ with the corresponding masks to obtain $\mathbf{F}^{final}$. (c) $\mathbf{F}^{final}$ and $w^{blend}$ are fed to the StyleGAN2 generator to synthesize the final output. (d) Blending loss consists of $\mathcal{L}^{keep}_{percept}, \lambda^{hair}_{percept}$, and $\mathcal{L}^{hair}_{percept}$.}
    \label{fig:blending}
\end{figure}
First, as presented in Fig.~\ref{fig:blending}(a), $w^{blend}$ is obtained by blending $w^{inpaint}_{src}$ and $w^{align}_{trg}$ with the blending weight $w^{weight}$.
$w^{blend}$ is formulated as $w^{inpaint}_{src} + w^{weight} \odot w^{align}_{trg}$, where $w^{weight}$ implies how much of $w^{align}_{trg}$ needs to be reflected to synthesize the final output.
Then, we prepare $\mathbf{F}$ tensors by feeding the first $m$ $w$ vectors to the pre-trained StyleGAN2 generator, as shown in Fig.~\ref{fig:blending}(b).
Here, we leverage $\mathbf{F}$ tensors in $FS$ space to effectively reconstruct the detailed spatial information~\cite{zhu2021barbershop} in the further blending.
We blend $\mathbf{F}^{align}_{trg}$, $\mathbf{F}^{blend}$, and $\mathbf{F}_{src}$ to gain $\mathbf{F}^{final}$ which contains detailed spatial information of the final output.
$\mathbf{F}^{final}$ is calculated as follows:
\begin{equation} 
    \mathbf{F}^{final} = \mathbf{M}_{trg\_hair}^{align} \odot \mathbf{F}^{align}_{trg} + \mathbf{M}^{blend} \odot \mathbf{F}^{blend} + \mathbf{M}^{keep}_{src} \odot \mathbf{F}_{src}.
\end{equation}
$\mathbf{F}^{align}_{trg}$ and $\mathbf{F}^{blend}$ are extracted from $w^{align}_{trg}$ and $w^{blend}$, respectively, and $\mathbf{F}_{src}$ is from the embedding step.
$\mathbf{M}_{trg\_hair}^{align}$ is a binary mask indicating the hair region in the aligned target hair image. $\mathbf{M}^{keep}_{src}$ is also a binary mask denoting the regions which are neither the source hair nor the aligned target hair. $\mathbf{M}^{keep}_{src}$ indicates the area that needs to be preserved in the source image.
Lastly, $\mathbf{M}^{blend}$ denotes the remaining regions.
The final output $\mathbf{\hat{I}}$ is generated from the pre-trained StyleGAN2 generator given $w^{blend}$ and $\mathbf{F}^{final}$ as inputs. Here, vectors in $w^{blend}$ except the first $m$ vectors are fed to the generator.

\noindent\textbf{Losses.}
In order to blend the previous optimized latent codes while preserving their structure and styles, we utilize the following losses.

First, to maintain the source face, clothes, background, etc., we apply the perceptual loss~\cite{zhang2018perceptual} on the valid regions in $\mathbf{S}^{keep}_{src}$ (\textit{i.e.}, the regions to be preserved in the source image) as follows:
\begin{equation}
    \mathcal{L}^{keep}_{percept} = \frac{1}{V}\sum_{i=1}^V \frac{1}{N_{\mathrm{VGG}^i}}\lVert \mathbf{M}^{keep}_{src} \odot ( \mathrm{VGG}^i(\mathbf{I}_{src}) - \mathrm{VGG}^i(\mathbf{\hat{I}}))\rVert_{1},
\end{equation}
where $\mathrm{VGG}^i$ denotes $i$-th layer of $\mathrm{VGG}16$ network~\cite{simonyan2014very} and $N_{\mathrm{VGG}^i}$ is the number of elements in the activation of $\mathrm{VGG}^{i}$.

Also, in order to preserve the aligned target hairstyle from the aligned target latent code $w^{align}_{trg}$, we use the hair perceptual loss formulated as follows:  
\begin{equation}
    \mathcal{L}^{hair}_{percept} = \frac{1}{V}\sum_{i=1}^V \frac{1}{N_{\mathrm{VGG}^i}}\lVert \mathbf{M}^{align}_{trg\_hair} \odot ( \mathrm{VGG}^i(\mathbf{I}^{align}_{trg}) - \mathrm{VGG}^i(\mathbf{\hat{I}}))\rVert_{1}.
\end{equation}

Lastly, we maintain the texture of the original target hair by utilizing the hairstyle loss $\mathcal{L}^{hair}_{style}$, where the style loss $\mathcal{L}_{style}$ is applied on the hair regions of the target hair image and the final output as $\mathcal{L}_{style}(\mathbf{M}_{trg\_hair}\odot\mathbf{I}_{trg},\mathbf{M}_{\hat{I}\_hair}\odot\mathbf{\hat{I}})$.

The total blending loss to optimize $w^{weight}$ is $\mathcal{L}^{keep}_{percept} + \lambda^{hair}_{percept}\mathcal{L}^{hair}_{percept} + \lambda^{hair}_{style}\mathcal{L}^{hair}_{style}$, where $\lambda^{hair}_{percept}$ and $\lambda^{hair}_{style}$ are the hyper-parameters to balance the relative importance between the losses.

\section{Experiments}

\subsection{Experimental Setup}
\textbf{Dataset.}
We utilize Flickr-Faces-HQ (FFHQ) dataset~\cite{karras2019stylegan} for hairstyle transfer and K-hairstyle~\cite{kim2021k} and VoxCeleb2~\cite{Chung18vox2} for reconstruction task.
For hairstyle transfer, we sample 6,000 pairs of two different identities (one for source and the other for target hairstyle) from 70,000 1,024$\times$1,024 images in FFHQ.

For the reconstruction, we create 500 test pairs by sampling the images from the K-hairstyle dataset, which includes 500,000 high-resolution multi-view images with more than 6,400 identities.
Following HairFIT~\cite{chung2021hairfit}, we filtered the images to remove the ones whose hairstyle is significantly occluded, or whose face is extremely rotated.
Additionally, we sample 500 pairs of a source and a target from more than 1 million videos in VoxCeleb2.
In the reconstruction task, two images in each pair have the same identity and different poses, and the source image in each pair is considered the ground truth image for the model to reconstruct. Each image is resized to 256$\times$256 in the experiments.

\noindent
\textbf{Baseline Models.}
We conduct a quantitative and qualitative comparison between our model and the following baselines: LOHO~\cite{saha2021loho}, Barbershop~\cite{zhu2021barbershop}, and HairFIT~\cite{chung2021hairfit}.
Here, we follow the official implementation code of LOHO and Barbershop.
Since LOHO utilizes an external inpainting network, we use a state-of-the-art inpainting network CoModGAN~\cite{zhao2021large}.
Also, we implement HairFIT with the codes and guidelines provided by the authors of HairFIT.


\subsection{Comparison to Baselines}
\textbf{Quantitative evaluations.}
First, we compare the fréchet inception distance (FID) score~\cite{heusel2017fid} of LOHO, Barbershop, and our model on hairstyle transfer task.
The FID score measures how similar the distributions of the synthesized images and the real images are, where the lower FID score indicates a higher similarity.
We compare 6,000 pairs of real and fake images, where each image is resized to 256$\times$256 for the evaluation.
As shown in the last column of Table~\ref{Table:posediff}, we achieve the lowest FID score compared to the baselines.

For further analysis, we compare the FID scores on three different levels of pose difference as conducted in the previous work~\cite{chung2021hairfit,saha2021loho}. 
We calculate the pose difference, PD, following the protocol presented in HairFIT~\cite{chung2021hairfit}.
In particular, we use 17 facial jaw keypoints extracted by the pre-trained 3D-keypoint extraction model~\cite{bulat2017far}.
The pose difference is calculated as $\mathrm{PD} = \frac{1}{17}\sum_{i=1}^{17}\rVert \mathbf{k}_{src}^i - \mathbf{k}_{trg}^i \rVert_{1}$, where $\mathbf{k}_{src}^i \in \mathbb{R}^{3}$ is a 3D coordinates of the $i$-th source keypoint and $\mathbf{k}_{trg}^i \in \mathbb{R}^{3}$ is a 3D coordinates of the $i$-th target keypoint.
Then, we divide the 6,000 pairs of a source and a target into three categories of 2,000 pairs: Easy, Medium, and Difficult.
As presented in Table~\ref{Table:posediff}, \ourmodel outperforms the other baselines for Medium and Difficult. 
Moreover, the margin between the FID scores of our model and other baselines increases as the pose difference increases from Easy to Difficult.

Additionally, we conduct a comparison with HairFIT on the reconstruction task using K-hairstyle and VoxCeleb2.
As in HairFIT, we measure the structural similarity (SSIM)~\cite{wang2004image} and learned perceptual image patch similarity (LPIPS)~\cite{zhang2018unreasonable} between generated images and ground truth images.
Table~\ref{Table:quant_hairfit} presents that our model outperforms HairFIT (except for LPIPS of VoxCeleb2) \textit{even without} learning to reconstruct different views of a source image using a multi-view dataset.

\begin{table}[!t]
    \centering
    \begin{tabular}{c|c|c|c|c} 
    \toprule
    Pose difference level & Easy & Medium & Difficult & Total \\
    \midrule
    LOHO~\cite{saha2021loho} & 21.70 & 23.40 & 28.36 & 19.63 \\ 
    Barbershop~\cite{zhu2021barbershop} & \textbf{20.75} & 21.45 & 26.30 & 18.07\\ 
    Ours & 20.79 & \textbf{20.56} & \textbf{22.72} & \textbf{17.06}\\ 
    \bottomrule
    \end{tabular}%
    \caption{Quantitative comparison with baselines. We measure the FID scores with three different levels of pose difference and with total pairs.} 
    \label{Table:posediff} 
\end{table}

\begin{table}[t!]
    \centering
    \begin{tabular}{c|cc|cc}
    \toprule
    Dataset & \multicolumn{2}{c|}{K-hairstyle} & \multicolumn{2}{c}{VoxCeleb2} \\ 
    \midrule
    Metric & SSIM$_{\uparrow}$ & LPIPS$_{\downarrow}$ & SSIM$_{\uparrow}$ & LPIPS$_{\downarrow}$\\
    \midrule
    HairFIT & 0.7242 & 0.2054 & 0.7520 & \textbf{0.2033} \\ 
    Ours & \textbf{0.7424} & \textbf{0.1786} & \textbf{0.7717} & 0.2078\\ 
    \bottomrule
    \end{tabular}
    \caption{Quantitative comparisons with HairFIT using multi-view datasets.
    }
    \label{Table:quant_hairfit}
\end{table}

\begin{figure*}[!t]
    \centering
    \includegraphics[width=0.82\linewidth]{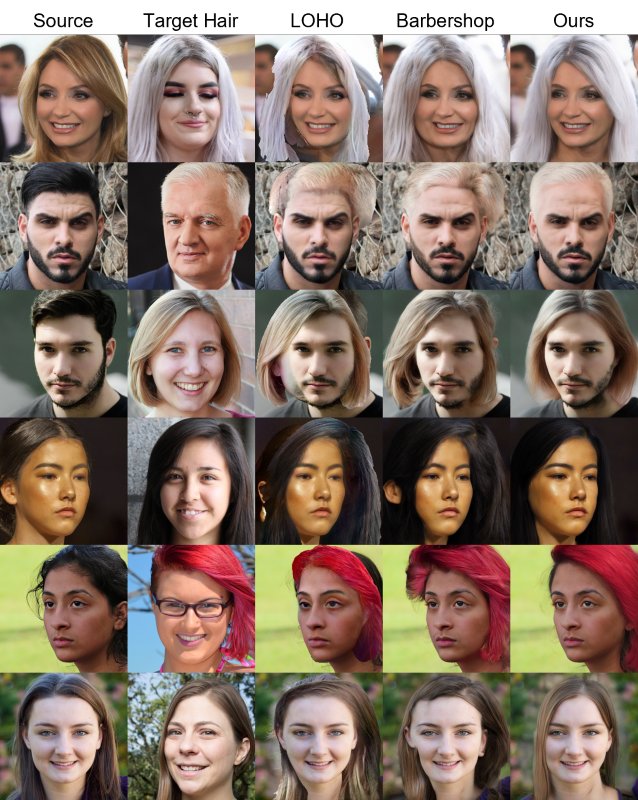}
    \caption{Qualitative comparison with the baselines on Difficult level of pose difference.}
    \label{fig:hard} 
\end{figure*} 

\begin{figure*}[!t]
    \centering
    \includegraphics[width=0.82\linewidth]{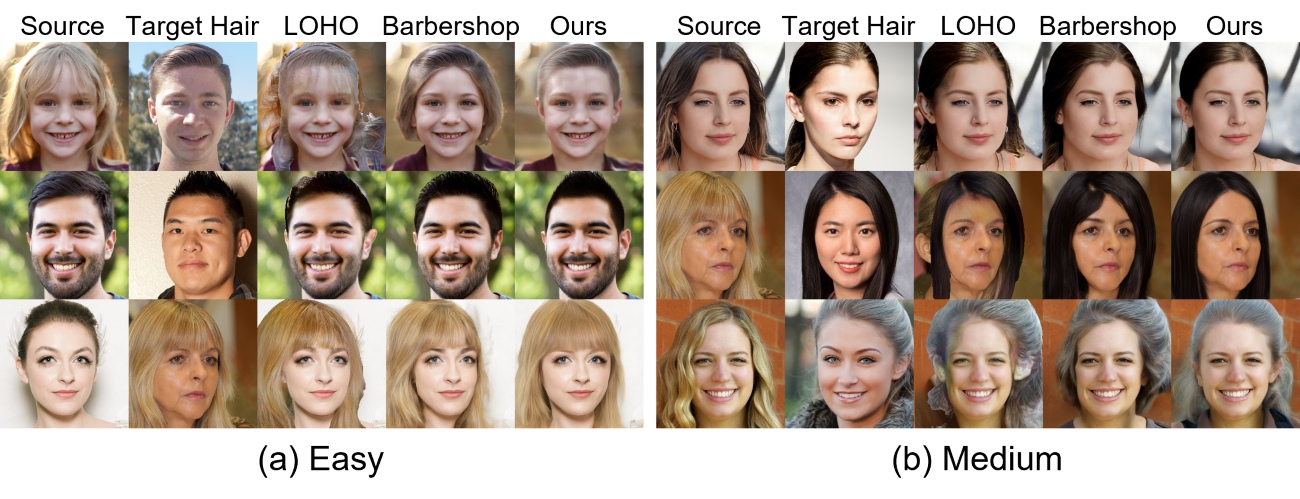}
    \caption{Qualitative comparison on (a) Easy and (b) Medium level of pose difference.} 
    \label{fig:easy} 
\end{figure*} 

\noindent\textbf{Qualitative evaluations.}
Fig.~\ref{fig:hard} and Fig.~\ref{fig:easy} demonstrate that \ourmodel successfully transfers the target hairstyle into the source regardless of the pose differences. 
Especially, as presented in Fig.~\ref{fig:hard}, \ourmodel shows superiority over other baselines on the Difficult level.
Furthermore, although the FID score of \ourmodel is slightly higher than Barbershop on Easy level, Fig.~\ref{fig:easy} present that the quality of \ourmodel is better to reflect the target hairstyle than the baselines.
The results show that \ourmodel successfully aligns the target hair to the source image, producing high-quality images of hairstyle transfer.
More results of the qualitative comparison are presented in the supplementary materials.

\begin{figure*}[!t]
    \centering
    \includegraphics[width=0.8\linewidth]{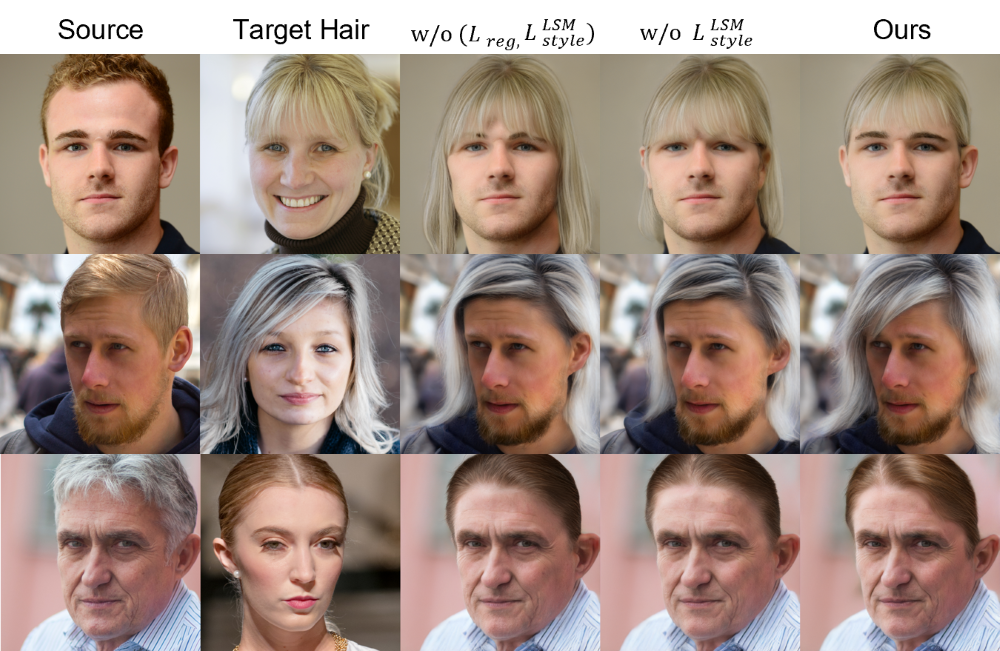}
    \caption{Qualitative ablation study on the losses in target hair alignment step.}  
    \label{fig:ablation} 
\end{figure*} 

\begin{table}[t]
    \centering
    \begin{tabular}{c|ccc}
    \midrule
    Configurations &  w/o ($\mathcal{L}_{reg}, \mathcal{L}^{LSM}_{style}$) & w/o $\mathcal{L}^{LSM}_{style}$ & Ours \\
    \midrule
    SSIM$_{\uparrow}$  & 0.7667 & 0.7716 & \textbf{0.7717}\\ 
    LPIPS$_{\downarrow}$  & 0.2125 & 0.2082 & \textbf{0.2078} \\ 
    \bottomrule
    \end{tabular}%
    \caption{Quantitative ablation study on the losses in target hair alignment step.}
    \label{Table:quant_abl}
\end{table}

\subsection{Ablation Study}
In the ablation study, we demonstrate the effectiveness of a \hairloss and regularization loss in our target hair alignment step.
We conduct a qualitative evaluation on hairstyle transfer using the FFHQ dataset and quantitative evaluation on the reconstruction task with VoxCeleb2.
In Fig.~\ref{fig:ablation} and Table~\ref{Table:quant_abl}, \textit{w/o} $(\mathcal{L}_{reg}, \mathcal{L}^{LSM}_{style})$ denotes our framework without $\mathcal{L}_{reg}$ and $\mathcal{L}^{LSM}_{style})$. Also, \textit{w/o} $\mathcal{L}^{LSM}_{style}$ indicates our framework without $\mathcal{L}^{LSM}_{style}$ and \textit{Ours} is our full framework.

The first row of Fig.~\ref{fig:ablation} indicates that the generated target hair is longer than the original target hair due to the absence of the $\mathcal{L}_{reg}$.
In the second row, the direction of the front hair of the outputs without $\mathcal{L}^{LSM}_{hair}$ are different from the original target hairstyle. 
Moreover, in the third row, the ``part" of the target hair is better reflected in the output of ours. 
The results present that our proposed losses effectively reflect the local style of the target hair while preserving its overall style. 
Additionally, as seen in Table~\ref{Table:quant_abl}, our full model outperforms other configurations with a gradual performance increase.
Although the difference between \textit{Ours} and \textit{w/o} $\mathcal{L}^{LSM}_{style}$ is marginal, the qualitative results presented above clearly illustrate the high visual quality of our full model in terms of preserving delicate hair features.
\begin{figure}[t!]
    \centering
    \includegraphics[width=0.9\linewidth]{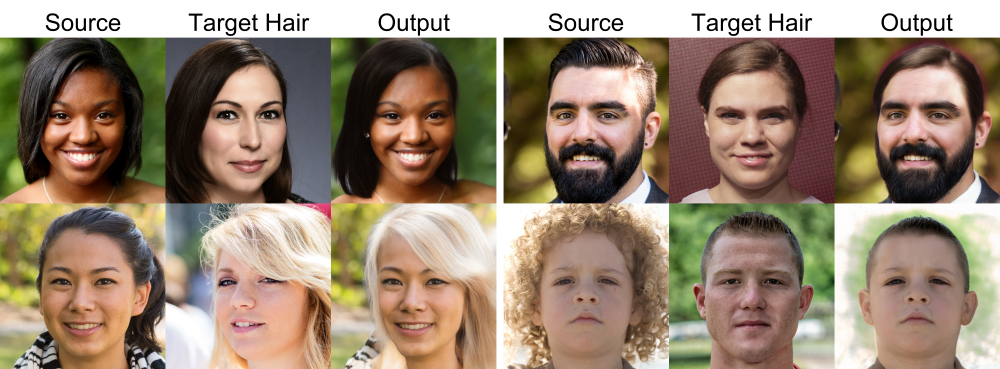}
    \caption{Limitations of our proposed method.}
    \label{fig:limit}
\end{figure}

\section{Discussions}
Although our model achieves a state-of-the-art performance compared to the baselines, several challenges still remain. 
First, since we transfer hairstyles via online latent optimization, it takes a few minutes on average for each image pair.
Also, our framework cannot newly generate the occluded part of the target hair due to the extremely turned head pose.
For example, the first three columns of Fig.~\ref{fig:limit} show that where the hair on the side is extremely occluded so that the final output barely has side hair.
Finally, the output might contain undesired background when the hair segmentation mask is inaccurately predicted.
The last three columns of Fig.~\ref{fig:limit} present undesirable background leaking.

\section{Conclusions}
This paper proposes a latent optimization framework for high-quality pose-invariant hairstyle transfer via local-style-aware hair alignment.
By leveraging latent optimization, we align the target hair without a multi-view dataset, while maintaining fine details of the hairstyle.
In addition, during the hair alignment, our newly-presented local-style-matching loss encourages our model to preserve the distinct structure and color of each local hair region in detail.
Finally, we perform occlusion inpainting and blending via latent optimization.
In this way, our model produces high-quality final output without noticeable artifacts.

\subsubsection{Acknowledgments.} This work was supported by the Institute of Information \& communications Technology Planning \& Evaluation (IITP) grant funded by the Korean government (MSIT) (No. 2019-0-00075, Artificial Intelligence Graduate School Program (KAIST) and the Ministry of Culture, Sports and Tourism and Korea Creative Content Agency (Project Number: R2021040097, Contribution Rate: 50).


\clearpage
%
%
\bibliographystyle{splncs04}
\bibliography{egbib}

\clearpage
\appendix

\noindent \textbf{\Large{Supplementary Material}}


\section{Qualitative Comparison with Additional Baselines}

\begin{figure}[h!]
    \centering
    \includegraphics[width=0.7\linewidth]{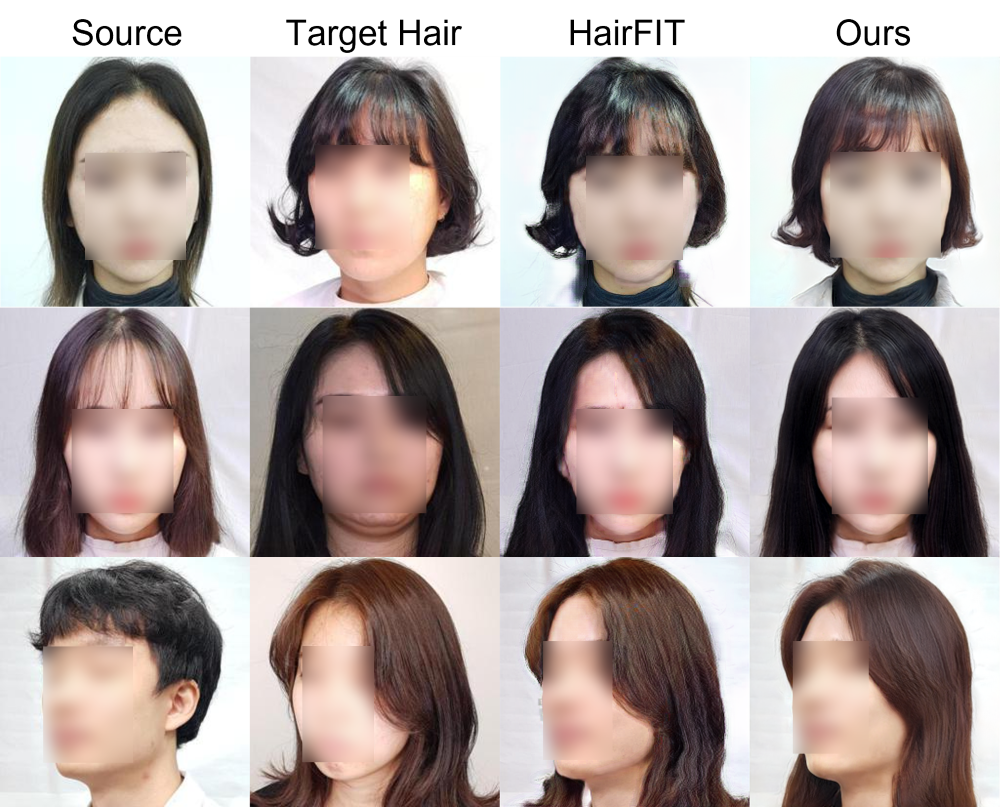}
    \caption{Qualitative comparison with HairFIT when a source and a target hair have similar poses. Note that we blur the face of the images from the K-hairstyle dataset due to the privacy issue.}
    \label{fig:hairfit-easy}
\end{figure}

As stated in our main paper, HairFIT~\cite{chung2021hairfit} proposes a pose-invariant hairstyle transfer model via flow-based hair warping and high-quality multi-view datasets.
Also, StyleFusion~\cite{kafri2021stylefusion} is a recently-proposed generative model which is capable of editing local features of an image (e.g., hairstyle in a facial image) by learning disentanglement of semantic regions in the StyleGAN~\cite{karras2019stylegan} latent space.
We conducted additional qualitative evaluation to demonstrate our superiority over HairFIT and StyleFusion.

\begin{figure}[t!]
    \centering
    \includegraphics[width=\linewidth]{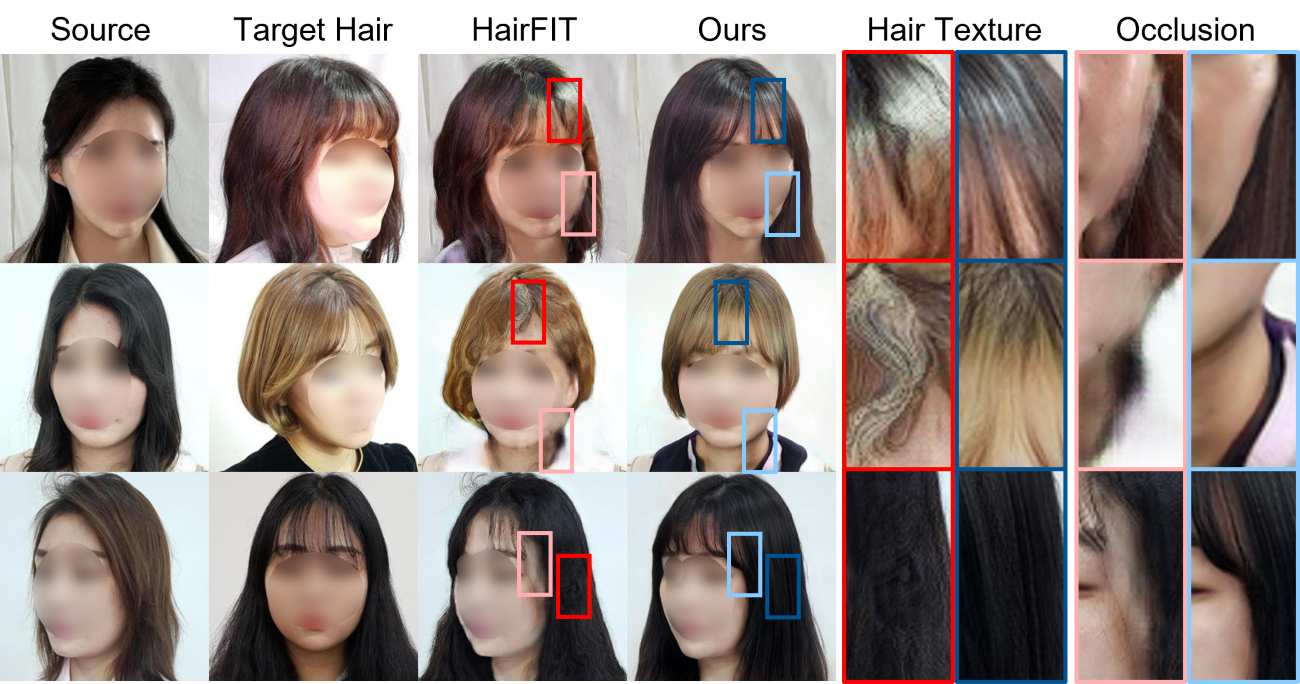}
    \caption{Qualitative comparison with HairFIT when a source and a target hair have different poses. The last two columns present zoomed-in regions of interest, each corresponding area indicated in the third and fourth columns. (Best viewed in color.) The second last column and the last column contain regions of hair texture and occluded regions in the source, respectively. Note that we blur the face of the images from the K-hairstyle dataset due to the privacy issue.}
    \label{fig:hairfit-hard}
\end{figure}

First, we compare our model with HairFIT.
We trained HairFIT in the same way described in the original paper and utilized the K-hairstyle dataset~\cite{kim2021k}.
The implementation codes and the dataset are provided by the authors of HairFIT.
K-hairstyle~\cite{kim2021k} includes 500,000 high-resolution multi-view hairstyle images with more than 6,400 identities.
Following HairFIT, we filtered the images to remove the ones whose hairstyle is significantly occluded, or whose face is extremely rotated.
The training set consists of 37,602 images with 4,291 identities, and the test set contains 4,309 images with 498 identities.
We cropped each image based on its hair and face segmentation mask and resized the images into $256\times256$ for a fair comparison.
For the embedding step of our framework, we trained StyleGAN2~\cite{karras2020stylegan2} with the same dataset before the inference.

Fig.~\ref{fig:hairfit-easy} and Fig.~\ref{fig:hairfit-hard} present the results with similar poses and with large pose differences, respectively.
Fig.~\ref{fig:hairfit-easy} illustrates that HairFIT achieves comparable performance to ours when a target hair is well aligned with a source image.
However, according to Fig.~\ref{fig:hairfit-hard}, HairFIT produces unrealistic outputs where a source and a target hair have different poses.
To be specific, HairFIT could not preserve the texture of straight strands of the target hairstyles, as shown in the fifth column of Fig.~\ref{fig:hairfit-hard}.
Moreover, HairFIT inpaints occluded regions, such as cheeks next to hair, neck, and shoulders, of a source image with undesirable blurry artifacts, as in the last column of Fig.~\ref{fig:hairfit-hard}.

\begin{figure}[t!]
  \centering
  \includegraphics[width=\linewidth]{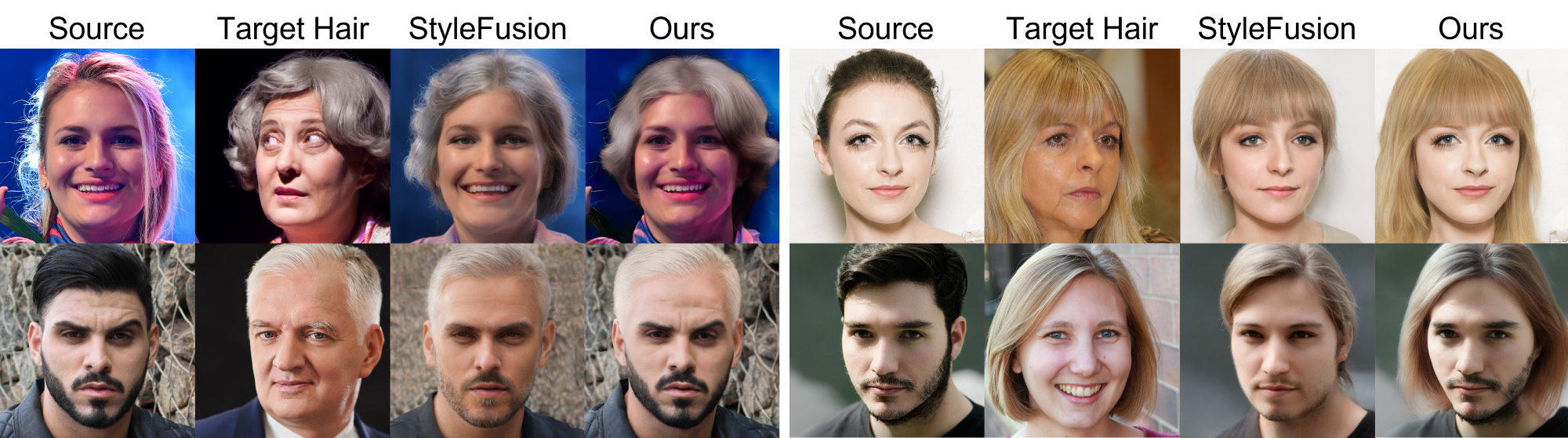}
  \caption{Qualitative comparison with StyleFusion.}
  \label{fig:stylefusion}
\end{figure}

Additionally, we conduct a qualitative comparison with StyleFusion.
We implemented the model with the official codes and utilized FFHQ dataset~\cite{karras2019stylegan} for the comparison.
Following the approach proposed in StyleFusion, we edit the `hair' attribute in the StyleGAN2 latent space to perform hairstyle transfer.
As in Fig.~\ref{fig:stylefusion}, StyleFusion is not shown to properly preserve the detailed textures as well as shapes of the target hairstyle. %
We speculate that the entangled attributes in the latent space (\textit{i.e.}, hair and inner face) prevent the model from producing fine details of the hair.

\begin{table}[t!]
    \centering
    \begin{tabular}{>{\centering}p{0.12\textwidth}|>{\centering}p{0.18\textwidth}|>{\centering}p{0.25\textwidth}|>{\centering}p{0.25\textwidth}||>{\centering\arraybackslash}p{0.10\textwidth}} 
    \toprule
    \makecell{ Ablated \\ Version} &
    \makecell{ Target Hair \\ Alignment }  & \makecell{Semantic Label of \\ Occlusion in $\mathbf{S}^{obj}_{src}$}  & \makecell{ $w^{inpaint}_{src}$ \\ Optimization } &  FID$_{\downarrow}$ \\
    \midrule
    (a) & \xmark & \xmark & in Blending & 43.37 \\
    (b) & \cmark & \xmark & in Blending & 39.68 \\
    (c) & \cmark & \cmark & in Blending & 32.69 \\ 
    Ours & \cmark & \cmark & in Source Inpainting & \textbf{18.02} \\
    \bottomrule
    \end{tabular}%
    \caption{Quantitative comparison with the ablated versions of our framework. (a), (b), and (c) indicate each ablated version, respectively.}
    \label{Table:sup-ablation} 
\end{table}

\begin{figure}[t!]
    \centering
    \includegraphics[width=1.0\linewidth]{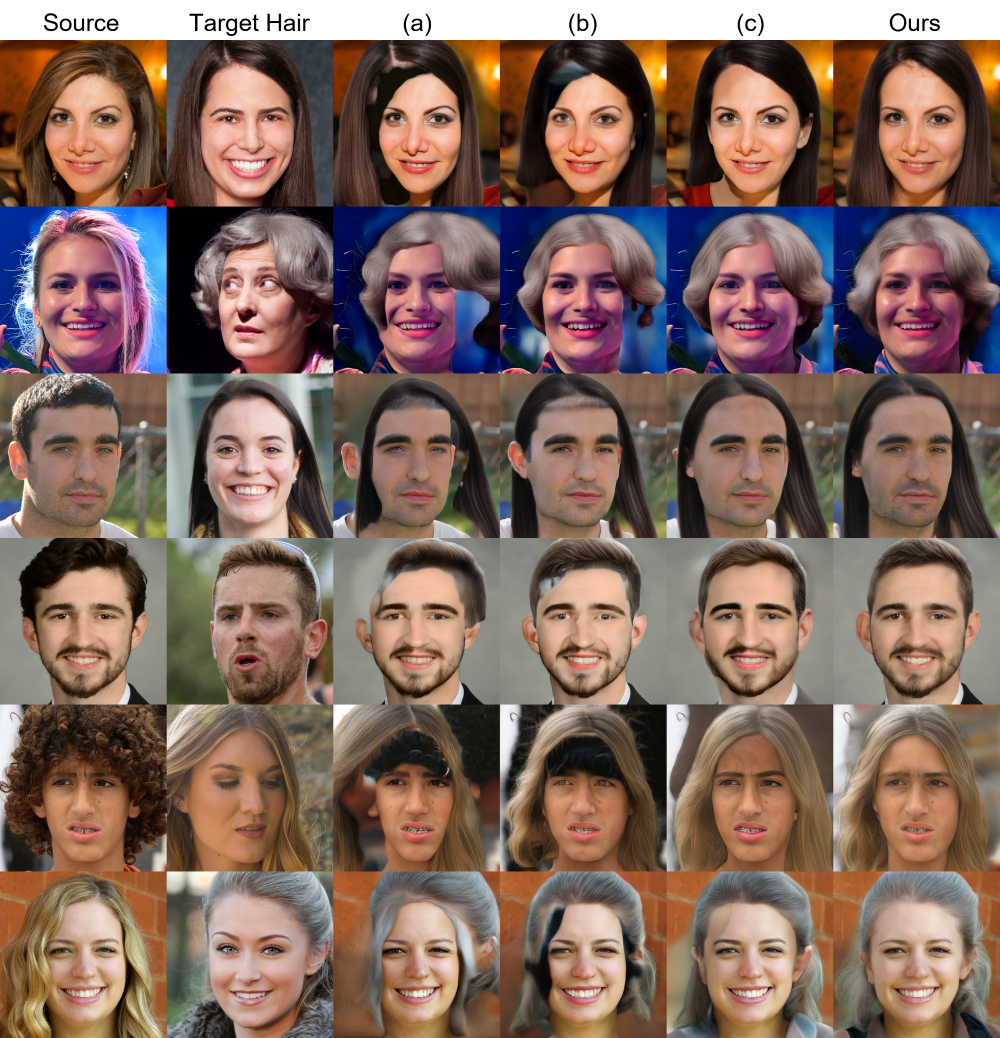}
    \caption{Qualitative comparison with the ablated versions of our framework using the FFHQ dataset. (a), (b), and (c) indicate each ablated version, respectively.}
    \label{fig:sup-ablation}
\end{figure}

\section{Additional Ablation Study}
To present the advantage of each step in our framework, we conduct additional quantitative and qualitative ablation studies using the FFHQ dataset.
For the quantitative evaluation, we measure the fréchet inception distance (FID) score~\cite{heusel2017fid}.

Starting from the embedding and blending step only, we gradually add each step to compare the corresponding results.
In Table~\ref{Table:sup-ablation} and Fig.~\ref{fig:sup-ablation}, we perform only the embedding and blending step in (a), append the target hair alignment step in (b), and add a semantic label of occluded regions to $\mathbf{S}^{obj}_{src}$ as a guide for the source inpainting in (c).
Lastly, in the last row, we include the source inpainting step, an independent optimization step for inpainting, which indicates our full framework.
Note that the source inpainting of (a), (b), and (c) is performed in the blending step, not in the independent source inpainting step.

According to Table~\ref{Table:sup-ablation}, the FID score gradually decreases as we add each step of our framework.
Since (a) does not have the target hair alignment step and a proper guide for the source inpainting, the corresponding outputs show dissatisfying quality.
The third column of Fig.~\ref{fig:sup-ablation} illustrates the results with misaligned hair and unrealistic occlusion inpainting.
Although (b) achieves the improved FID score with the aid of the target hair alignment step, source occlusions of the outputs are filled with unnatural textures, as presented in the fourth column of Fig.~\ref{fig:sup-ablation}. 
Since a lack of semantic label of occluded regions in $\mathbf{S}^{obj}_{src}$ cannot provide a proper guide for the source inpainting, (b) allows the occluded regions to be inpainted with random undesirable textures.
On the other hand, (c) produces the results with advanced quality, especially in the regions of source occlusion, with an appropriate assist of $\mathbf{S}^{obj}_{src}$.
However, the fifth column of Fig.~\ref{fig:sup-ablation} indicates that the final outputs include regions inpainted with undesirable colors or textures.
This is because the source inpainting, \textit{i.e.}, the optimization of $w^{inpaint}_{src}$, is conducted simultaneously with the optimization of a blending weight $w^{weight}$ in blending step.

To address this issue, we added an independent $w^{inpaint}_{src}$ optimization step only for source inpainting in our final framework, which achieves superior performance both quantitatively and qualitatively.
The last column of Fig.~\ref{fig:sup-ablation} presents the results of our full framework with a superior quality of target hair alignment and occlusion inpainting compared to other configurations.

\begin{figure}[t]
  \centering
  \includegraphics[width=\linewidth]{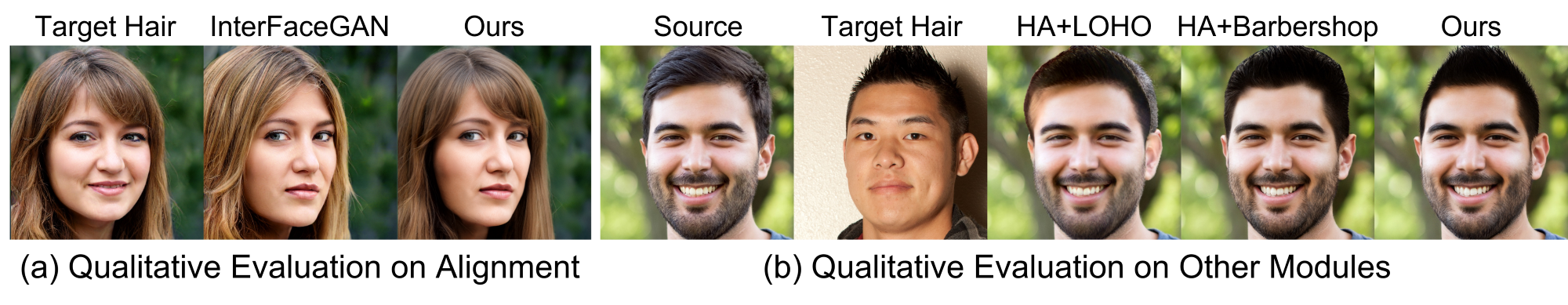}
  \caption{(a) Visual comparison with InterFaceGAN on the alignment module and (b) qualitative comparison with baselines equipped with our alignment module.}
  \label{fig:qualitative}
\end{figure}

\section{Analysis of Individual Modules}
In this section, we further analyze each module of our framework.
First, we conduct an additional evaluation on the target hair alignment module (HA).
As a baseline, we adopt InterFaceGAN~\cite{shen2020interpreting} which has the capability of aligning the target pose similar to a source via latent vector interpolation.
Fig.~\ref{fig:qualitative}(a) demonstrates the qualitative result on alignment, where the target hair is manipulated to show the same objective pose.
The objective pose is selected by randomly interpolating the target hair along the pose boundary of InterFaceGAN.
InterFaceGAN shows high performance on pose alignment but inappropriately alters the target hairstyle.
In contrast, our model properly produces a pose-aligned image while maintaining the target hair details.

Next, we evaluate the rest of our modules except for HA, by combining HA with the baselines: LOHO and Barbershop.
To this end, we perform a user study with 20 graduate students, to compare 20 images generated by three different configurations: HA + LOHO, HA + Barbershop, and ours.
For each pair, a participant is asked to select a top-1 sample with two criterion: (1) preservation of delicate features of a target hair and (2) inpainting quality against source occlusion.
The study results show that 71\% and 63\% of our method results were selected as the top-1 sample for each criteria, respectively.
Fig.~\ref{fig:qualitative}(b) shows qualitative examples appeared in the user study.
As can be seen, our model visually outperforms the baselines by successfully preserving the texture and shape of a target hair.

\section{Implementation Details}
The optimization step size of embedding $W+$, embedding $FS$, target hair alignment, source inpainting, and blending are 1,100, 250, 100, 140, and 400, respectively.
In the target hair alignment and blending step, we set all lambdas of the losses as 1.

The optimization is conducted on a single GeForce RTX 3090 GPU and it requires 10GB GPU memory.
For the inference time, the embedding step takes less than 2 minutes per image, and all the other steps take 78 seconds on average in total.

\begin{figure}[t!]
    \centering
    \includegraphics[width=0.95\linewidth]{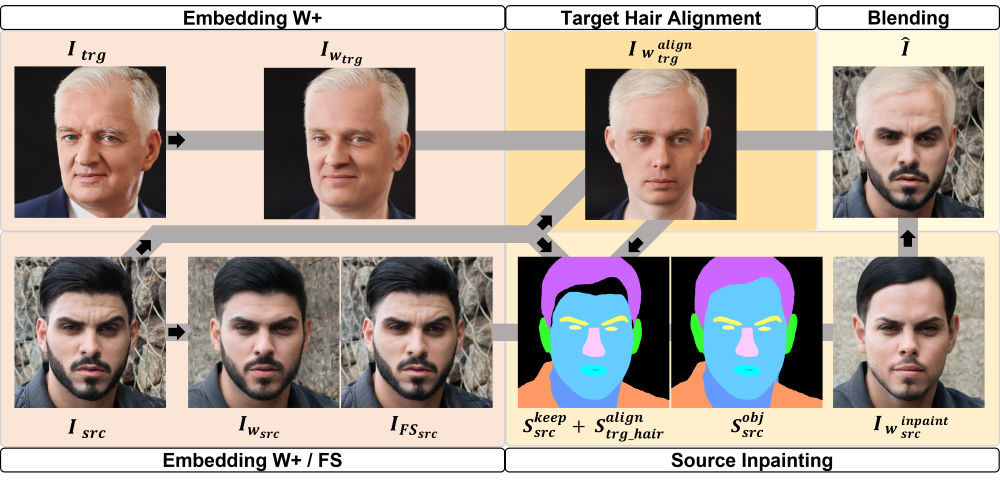}
    \caption{Visualization of results from each step in our framework.}
    \label{fig:sup-overview}
\end{figure}

\begin{figure}[t!]
    \centering
    \includegraphics[width=0.95\linewidth]{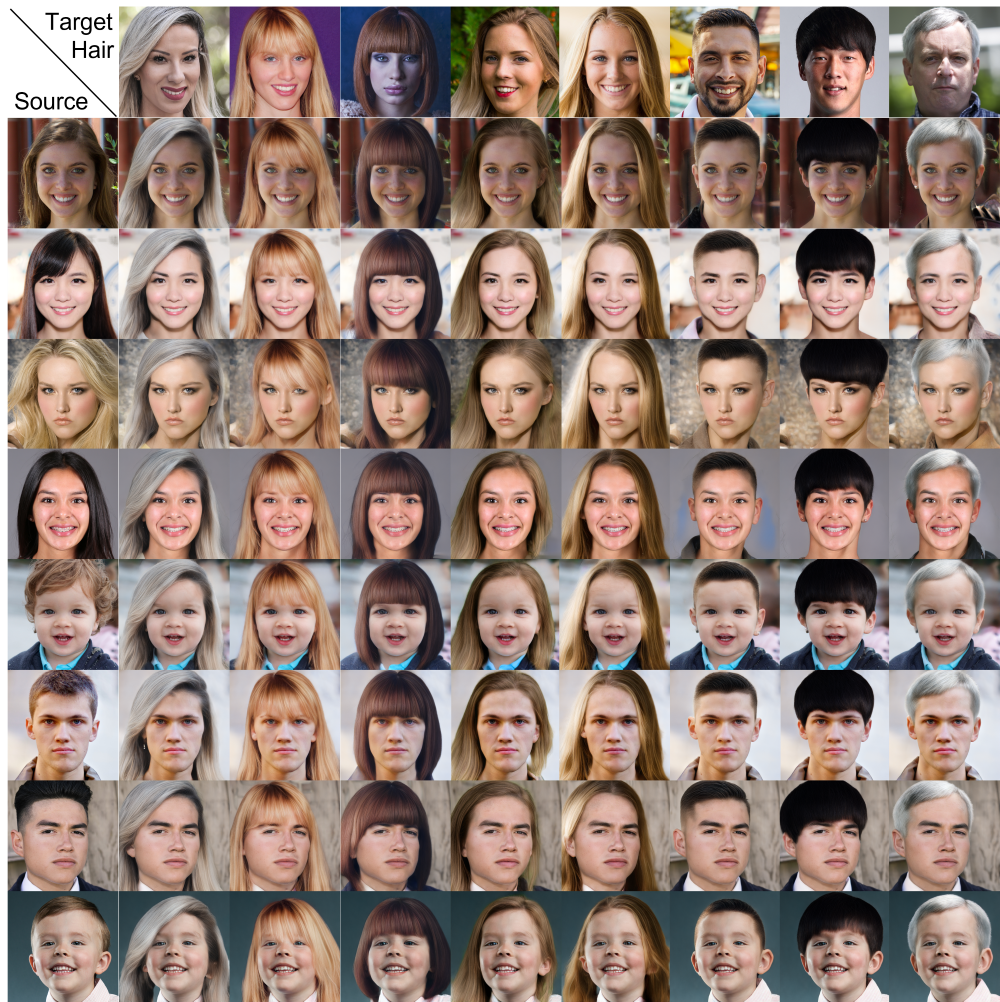}
    \caption{Additional qualitative results of our framework with the FFHQ dataset.}
    \label{fig:sup-qualitative}
\end{figure}

\section{Additional Qualitative Results}
First, to further understand our framework, we visualize an example qualitative result with its intermediate outputs from each step in Fig.~\ref{fig:sup-overview}.
To be specific, given the source image $I_{src}$ and the target hair image $I_{trg}$, $I_{w_{trg}}$, $I_{w_{src}}$, and $I_{{FS}_{src}}$ are the images reconstructed from the embedded latent codes $w_{trg}$, $w_{src}$, and ${FS}_{src}$ obtained in the embedding step. 
Also, $I_{w^{align}_{trg}}$ is the aligned target hair image generated from $w^{align}_{trg}$ obtained in the target hair alignment step.
Then, in the source inpainting step, we first create an objective label $S^{obj}_{src}$ for source inpainting based on $S^{keep}_{src}$ from $I_{src}$ and $S^{align}_{trg_hair}$ from $I_{w^{align}_{trg}}$. 
By optimizing $w_{src}$ to follow $S^{obj}_{src}$, we obtain inpainted source latent code $w^{inpaint}_{src}$, which is visualized in $I_{w^{inpaint}_{src}}$.
Finally, the final output $\hat{I}$ is generated via the blending step, where we blend $w^{align}_{trg}$ and other features in $w_{src}$ and $w^{inpaint}_{src}$.

Additionally, Fig.~\ref{fig:sup-qualitative} presents additional qualitative results with FFHQ dataset.
We transfer various target hairstyles on the first row of Fig.~\ref{fig:sup-qualitative} to each of the source images in the first column of Fig.~\ref{fig:sup-qualitative}.

\end{document}